\documentclass[11pt]{article}
\usepackage{coling2018}
\usepackage{times}
\usepackage{url}
\usepackage{latexsym}

\usepackage{graphicx}
\usepackage{mathtools}
\usepackage{tabularx}
\usepackage{multirow,bigdelim}
\usepackage{booktabs}

\usepackage{url}

\usepackage{xargs}
\usepackage{graphicx}
\usepackage[pdftex,dvipsnames]{xcolor}
\usepackage[colorinlistoftodos,prependcaption,textsize=tiny]{todonotes}

\newcommandx{\tdAh}[2][1=]{\todo[linecolor=Blue,backgroundcolor=Blue!25,bordercolor=Blue,#1]{Ahm: #2}}

\title{Assessing Composition in Sentence Vector Representations}

\author{Allyson Ettinger$^1$, Ahmed Elgohary$^2$, Colin Phillips$^1$, Philip Resnik$^{1,3}$ \\
  $^1$Linguistics, $^2$Computer Science, $^3$Institute for Advanced Computer Studies \\
  University of Maryland, College Park, MD \\
  {\tt \{aetting,colin,resnik\}@umd.edu, elgohary@cs.umd.edu} }
  
\date{}

\begin{document}
\maketitle
\begin{abstract}

An important component of achieving language understanding is mastering the composition of sentence meaning, but an immediate challenge to solving this problem is the opacity of sentence vector representations produced by current neural sentence composition models. We present a method to address this challenge, developing tasks that directly target compositional meaning information in sentence vector representations with a high degree of precision and control. To enable the creation of these controlled tasks, we introduce a specialized sentence generation system that produces large, annotated sentence sets meeting specified syntactic, semantic and lexical constraints. We describe the details of the method and generation system, and then present results of experiments applying our method to probe for compositional information in embeddings from a number of existing sentence composition models. We find that the method is able to extract useful information about the differing capacities of these models, and we discuss the implications of our results with respect to these systems' capturing of sentence information. We make available for public use the datasets used for these experiments, as well as the generation system.\footnote{Code for the generation system, as well as a pointer to the classification datasets, can be found at\\ \url{https://github.com/aetting/compeval-generation-system}} 
\end{abstract}

%
%
\blfootnote{
    %
    %
    \hspace{-0.65cm}  
    %
    
    %
  
    \hspace{-0.65cm}  
    This work is licensed under a Creative Commons 
    Attribution 4.0 International License.
    License details:
    \url{http://creativecommons.org/licenses/by/4.0/}
}

\section{Introduction}




As natural language processing strives toward language understanding, it is important that we develop models able to extract and represent the \emph{meaning} of sentences. Such representations promise to be applicable across a variety of tasks, and to be more robust than non-meaning-based representations for any given task requiring meaning understanding. To accomplish meaning extraction, a particular need is that of mastering composition: systematic derivation of the meaning of a sentence based on its parts. 

In this paper we tackle compositional meaning extraction by first addressing the challenge of evaluation and interpretability: after all, in order to improve meaning extraction, we need to be able to evaluate it. But with sentence representations increasingly taking the form of dense vectors (embeddings) from neural network models, it is difficult to assess what information these representations are capturing---and this problem is particularly acute for assessing abstract content like compositional meaning information. 

Here we introduce an analysis method for targeting and evaluating compositional meaning information in sentence embeddings. The approach builds on a proposal outlined in \newcite{ettinger2016probing}, and involves designing classification tasks that directly target the information of interest (e.g., ``Given a noun $n$, verb $v$, and an embedding $\mathbf{s}$ of sentence $s$: is $n$ the \emph{agent} of $v$ in $s$?''). By contrast to related work analyzing surface variables like word content and word order in sentence embeddings \cite{adi2016fine}, we specifically target compositional meaning information relevant to achieving language understanding---and in order to isolate this more abstract information, we exert careful control over our classification datasets to ensure that we are targeting information arrived at by composition of the source sentence, rather than general statistical regularities. Our approach is informed by methods in cognitive neuroscience and psycholinguistics, where such controls are standard practice for studying the brain.

In particular, to ensure validity of our tests we introduce three mechanisms of control. First, to create controlled datasets at the necessary scale, we develop a generation system that allows us to produce large sentence sets meeting specified semantic, syntactic and lexical constraints, with gold-standard meaning annotation for each sentence. Second, we control the train-test split so as to require more robust generalization in order to perform the tasks successfully. Third, we employ a sanity check leveraging known limitations of bag-of-words (BOW) composition models: for any tasks requiring order information \emph{from the source sentence}, which BOW models cannot logically retain, we check to ensure that BOW composition models are at chance performance.

These controls serve to combat a problem that has gained increasing attention in recent work: many existing evaluation datasets contain biases that allow for high performance based on superficial cues, thus inflating the perceived success of systems on these downstream tasks~\cite{gururangan2018annotation,bentivogli2016sick}. In the present work, our first priority is careful control of our tasks such that biases are eliminated to the greatest extent possible, allowing more confident conclusions about systems' compositional capacities than are possible with existing metrics. 

The contributions of this paper are threefold. 1) We introduce a method for analyzing compositional meaning information in sentence embeddings, along with a generation system that enables controlled creation of datasets for this analysis. 2) We provide experiments with a range of sentence composition models, to demonstrate the capacity of our method to shed light on compositional information captured by these models. 3) We make available the classification datasets used for these experiments, as well as the generation system used to produce the sentence sets, to allow for broader testing of composition models and to facilitate creation of new tasks and classification datasets.

Although we focus on neural composition models and sentence embeddings in the present paper---due to the current dominance of these methods and the need to evaluate their compositional capacities---it is important to note that this analysis method can also be applied more broadly. Since the method simply operates by classification of sentence representations, it can be applied to any format of sentence representation that can be input as features to a classifier.

\section{Meaning and composition}\label{sec:meancomp}
In this section we will briefly explain the concepts of \emph{meaning} and \emph{composition}, which are the central targets of our analyses in this work.

Our approach assumes there to be identifiable components of \emph{meaning} that we can expect in well-formed sentence representations. For instance, the sentence ``the dog chased the girl'' contains the information that there was a chasing event, and a dog was the chaser (\emph{agent} of chasing) and a girl the chasee (\emph{patient} of chasing). The sentence ``the dog did not bark'' conveys that a barking event did not happen. 

Humans are able to extract meaning with remarkable robustness, and a key factor in human language understanding is \emph{composition}: the productive combinatory capacity that allows sentence meaning to be derived systematically based on the meanings of its parts~\cite{heim1998semantics}. To illustrate the power of this systematicity, consider a nonsensical sentence like the following:

\begin{center}
\emph{The turquoise giraffe recited the sonnet but did not forgive the flight attendant.} \\
\end{center}
Though this sentence describes an entirely implausible scenario, and though nothing like it should ever have occurred in any corpus or conversation, any English speaker is able to extract the meaning of this sentence without difficulty. This is because language is highly systematic, and the meanings of the parts of the sentence can be combined predictably to arrive at the full meaning. 

Regardless of how closely NLP systems should draw on human strategies for language processing, the need for composition is clear: if systems do not construct meanings of sentences based on their parts, then the alternative is memorization of all possible sentences, which is neither practical nor possible.

In this work, critically, we are focused on the results of systematic compositional processes, to be distinguished from biases based on general statistical regularities. The importance of this distinction is highlighted by the result reported in \newcite{adi2016fine}, which shows a BOW composition model attaining 70\% accuracy on a binary word order classification task. This result is surprising given that BOW models (which simply average together word-level representations) necessarily sacrifice any order information from the source sentence. This suggests that the above-chance performance relies on statistical regularities of word ordering in the data as a whole, independent of the source sentence---that is, the model's above-chance performance must be dependent on some correspondence between word orders being tested and word orders seen when training the word embeddings.

Although sensitivity to such regularities is often useful, in this work we are concerned with \emph{systematic composition of the source sentence itself}, abstracting away from general statistical regularities. This is critical for our purposes: to master composition, models must be able to construct the meaning of a sentence not only when it matches commonly-seen patterns (e.g., ``the cat chased the mouse'') but also when it deviates from such patterns (e.g., ``the mouse chased the cat''). This is the reasoning behind our BOW sanity check, discussed in Section~\ref{sec:bow}, which serves to ensure that our tests cannot be solved by simple averaging. Additionally, the biases in naturally-occurring data, further highlighted by the Adi et al. result, motivate our use of generated data for the sake of maintaining the necessary level of control.

\section{The present method}\label{sec:method}

\subsection{Approach}

The approach that we take to probe for compositional meaning information in sentence embeddings is inspired by the neuroscience technique of multivariate pattern analysis~\cite{haxby2014decoding}, which tests for encoding of information in patterns of neural data by means of classification tasks designed to be contingent on the information of interest. Our use of careful control in implementing this approach is also informed more generally by the methodologies of cognitive neuroscience and psycholinguistics, which standardly use these kinds of controls to draw conclusions about information in human brain activity. The approach that we develop here builds on the proposal of \newcite{ettinger2016probing}---which described the basic form of the method and provided a simple validation with a small set of active and passive sentences. In the present work we flesh out and strengthen the method with a number of more rigorous controls aimed at better isolating the information of interest, and we substantially expand the scope of the tests through the use of a more sophisticated sentence generation system. 


\subsection{Classification tasks} 

As proposed by~\newcite{ettinger2016probing}, we target two meaning components as our starting point: semantic role and negation. These components are priorities because they are fundamental to the meaning of a sentence, having bearing on the key questions of ``what happened (and what didn't)'' and ``who did what to whom''. Additionally, they represent information types that can be heavily distorted with respect to surface variables like word content and order: to know semantic role and negation information, it is not enough to know which words are in the sentence or which words come earlier in the sentence.  

We formulate the semantic role classification task (``\textbf{SemRole}'') as follows: ``Given representation $\mathbf{n}$ of probe noun $n$, representation $\mathbf{v}$ of probe verb $v$, and embedding $\mathbf{s}$ of sentence $s$ (with $s$ containing both $n$ and $v$), does $n$ stand in the AGENT relation to $v$ in $s$?'' For example, an input of \{$n$: ``professor'', $v$: ``help'', $s$: ``the professor helped the student''\} would receive a positive label because \emph{professor} is AGENT of \emph{help} in the given sentence.

We formulate the negation classification task (``\textbf{Negation}'') as follows: ``Given a representation $\mathbf{v}$ of a probe verb $v$, and an embedding $\mathbf{s}$ of sentence $s$ (with $s$ containing $v$, one negation, and one other verb), is $v$ positive or negated in $s$?'' For example, an input of \{$v$: ``sleep'', $s$: ``the professor is not actually helping the student who is totally sleeping''\} receives a positive label because \emph{sleep} is not negated in that sentence. To decouple this from a simpler task of identifying adjacency between negation and a verb, we insert variable-length adverb sequences (e.g., \emph{not \textbf{really, actually} helping}) before the verbs in the dataset (negated and non-negated), to ensure that the negation is not always adjacent to the verb that it affects.

These formulations differ from those in the original~\newcite{ettinger2016probing} proposal, instead making use of variable word probes as employed by \newcite{adi2016fine}. This adjustment was made to maximize the generalization required for strong performance on the tasks, and to further reduce vulnerability to biasing cues in the datasets. More detail on our implementation of this formulation is given in Section~\ref{sec:probes}.

\subsection{Means of control}

The most critical consideration in this work is ensuring that we can draw valid conclusions about composition from performance on our classification tasks. To this end, we take a number of measures to control our data, to avoid biasing cues that would make the tasks solvable independent of the information of interest---a problem observed in many existing datasets, as mentioned above~\cite{gururangan2018annotation}.

\paragraph{Generation system}
A critical component of isolating abstract meaning information is employing syntactic variation, such that the meaning information of interest is the single underlying variable distinguishing label categories. For instance, we might use sentences like ``the professor helped the student'', ``the student was helped by the professor'', and ``the student that the professor helped was sleeping''---which vary in structure, but which share an underlying event of a professor helping a student.

In order to produce sentence sets that exhibit this level of variation---and that reach the necessary scale for training and testing classifiers---without allowing the statistical biases of naturally-occurring data, we developed a generation system that takes as input lexical, semantic and syntactic constraints, and that produces large sentence sets meeting those constraints. In addition to allowing us to produce controlled datasets, this system also ensures that the generated datasets are annotated with detailed semantic and syntactic information. This generation system is described in greater detail in Section~\ref{sec:gensys}.  

\paragraph{Train/test splits}

To be confident that the classifier is picking up on underlying meaning information and not simply a union of different superficial cues across syntactic structures, we make careful provisions in our train/test split to ensure generalization (beyond the obvious split such that sentences in test do not appear in training).  For our semantic role task, certain ($n$,$v$) probe combinations are held out for test, such that no combinations seen at test time have been seen during training. This is done to ensure that the classifier cannot rely on memorized sequences of words. For our negation task, which uses only one probe, we hold out certain adverbs from training (as described above, adverbs are used as material to separate the negation and the verb), such that at test time, the material separating the negation and the verb (or preceding the non-negated verb) has never been seen in training.

\paragraph{BOW as control}\label{sec:bow}

As described above, it is logically impossible for BOW models to encode information that requires access to word order from the source sentence itself. We leverage this knowledge to create a sanity check baseline for use in monitoring for lexical biases: if, for any task requiring access to word order information, the BOW baseline performs above chance, we know that the datasets contain lexical biases affecting the classification results, and we can modify them accordingly. 

\section{Generation system}\label{sec:gensys}

In this section we describe the generation system that we use to create large, controlled datasets for our classification tasks. As described above, this system takes input constraints targeting semantic, syntactic, and lexical components, and produces diverse, meaning-annotated sentences meeting those constraints.

\subsection{Event/sentence representations}
\begin{figure*}[ht]
\begin{center}
\includegraphics[width =.6\textwidth]{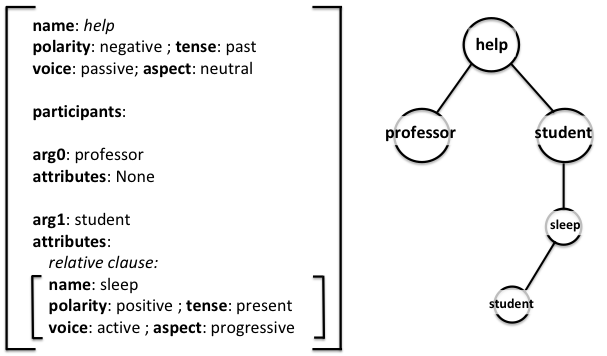}
\end{center}
\caption{Event representation for \emph{``The student who is sleeping was not helped by the professor''}}\label{evrep}
\end{figure*}

As a framework for specifying semantic and syntactic constraints, we use a class of event representations that contain both lexicalized semantic information and necessary syntactic information, such that there is a deterministic mapping from a fully-populated event representation to a corresponding surface sentence form.  These representations fall roughly within the category of ``lexicalized case frame'' outlined by Reiter and Dale~\shortcite{reiter2000building} for natural language generation. Figure~\ref{evrep} shows an example representation, in fully-specified textual form, and in simplified graphical form.  

Our representations are currently restricted to events denoted by transitive and intransitive verbs, with the arguments of those verbs and optional transitive or intransitive relative clauses on those arguments.

These representations are comparable in many ways to abstract meaning representation (AMR) \cite{banarescu2012abstract}, but rather than abstracting entirely away from syntactic structure as in AMR, our event representations encode syntactic information directly, along with the more abstract meaning information, in order to maintain a deterministic mapping to surface forms. Relatedly, while AMR uses PropBank frames~\cite{palmer2005proposition} to encode meaning information, we encode information via English lemmas, to maintain control over lexical selection during generation. 

These representations can be partially specified to reflect a desired constraint, and can then be passed in this partial form as input to the generation system---either as a required component, or as a prohibited component. This allows us to constrain the semantic and syntactic characteristics of the output sentences. In addition to partial events, the system can also take lists of required or prohibited lexical items.

\subsection{Event population}

The system uses a number of structural templates into which partial events can be inserted. Structural templates vary based on the transitivity of verbs and the presence or absence of relative clauses on arguments---for instance, if the nodes in the right side of Figure~\ref{evrep} were unpopulated, it would depict an empty structural template consisting of a transitive main verb with an intransitive relative clause on arg1. Once we have inserted a partial event into a subsection of an empty structural template (events can be inserted into either the main clause or a relative clause), the system populates the remainder of the event components by iterating through available verbs and nouns of the vocabulary, and through available values for unfilled syntactic characteristics (such as polarity, tense, voice, etc.). 

For simplicity, we control plausibility of argument/predicate combinations by setting the system vocabulary such that it contains only animate human nouns, and only verbs that can take any of those nouns in the relevant argument slots. This is a reasonable task due to the capacity of the system to generate thousands of sentences from only a handful of nouns and verbs. We leave incorporation of more sophisticated selectional preference methods~\cite{van2014neural,resnik1996selectional} for future work.

Our goal is to find the optimal balance between the critical need of this method for structurally variable, carefully controlled sentences, and the practical need to avoid substantial deviation from sentence types to which systems will have been exposed during training. To this end, we draw our vocabulary from comparatively frequent words, and we impose structural constraints to limit the complexity of sentences---specifically, in the current experiments we restrict to sentences with no more than one relative clause, by omitting templates that include relative clauses on both arguments of a main verb.

\subsection{Syntactic realization}
Once an event representation is fully populated, it is submitted to a surface realization module that maps from the event to a surface sentence via a simple rule-based mapping. Since the representations specify syntactic information and use lexicalized meaning information, there is no significant process of lexical selection required during surface realization---only morphological inflection derivable from syntactic characteristics. As a result, the event representations map deterministically to their corresponding surface forms.  We use a grammar specified using the NLTK feature grammar framework ~\cite{bird2009natural}. Morphological inflections are drawn from the XTAG morphological database~\cite{doran1994xtag}.

\subsection{Sentence quality}

To ensure the quality of generation system output, we manually inspected large samples of generated sentences throughout development and after generation of the final sets, to confirm that sentences were grammatical and of the expected form. Table~\ref{tab:sentences} shows a representative sample of generated sentences.\footnote{More sentences can be found in the publicly available classification datasets.} 

\begin{table*}[ht!]
  \centering
\begin{tabular}{ l } \toprule
the men were sleeping \\
the woman followed the lawyer that the student is meeting \\ 
the women were being helped by the lawyers \\ 
the student called the man \\
the scientist that the professors met is dancing \\ 
the doctors that helped the lawyers are being recommended by the student \\ \bottomrule
\end{tabular}
  \caption{Example generated sentences}
  \label{tab:sentences}
\end{table*}

\section{Implementation of lexical variability}\label{sec:probes}
As discussed above, we adopt the variable probe formulation used by \newcite{adi2016fine}. This adds a dimension to the learning task that is not present in the original~\newcite{ettinger2016probing} task formulation: the classifier needs not only to identify meaning information in the input sentence---it needs to identify meaning information contingent on the identities of the particular probe words. 

To identify the probe word(s) in the input features,~\newcite{adi2016fine} use the source word embeddings, but this is problematic for our purposes, given that we want to test a wide variety of models, which use word embeddings of different types and sizes. To avoid this variability, it would be preferable to use one-hot vectors to identify word probes. To this end, we performed a series of experiments testing whether classification accuracy was affected by use of one-hot probe representations by comparison to embedding probes, in a replication of the word content task of~\newcite{adi2016fine}. Finding almost equivalent accuracies between the two input types, we use one-hot probe representations in all subsequent experiments. 

Note that as a result, by contrast to~\newcite{adi2016fine} we are not assuming the classifier to identify words in the sentence representation based on resemblance to their original word embeddings---this may not in fact be a safe assumption, given that the word's representation may distort during composition of the sentence. Instead, the classifier must learn a mapping from each one-hot representation to its manifestation in sentences. This means that all words must appear as probes in training. To facilitate the learning of this mapping, we restrict to a small (14-lemma) vocabulary of noun and verb probes in these experiments.\footnote{Sentences themselves contain various morphological inflections of these lemmas.} Because the generation system is able to produce thousands of sentences from even such a restricted vocabulary as this, this limitation does not prevent generation of adequately large datasets. 

A note about the size and selection of the vocabulary: some composition tests will surely be sensitive to specific idiosyncrasies of individual words, in which case the choice of vocabulary will be of great importance. However, for the particular semantic role and negation tasks described here, the focus is on identification of structural dependencies between words, which are not in this case sensitive to the specific nouns/verbs used. Consequently, for these tasks---as long as vocabulary words are not out-of-vocabulary for the models (which we confirm below)---the important thing should be not what the words themselves are, but whether dependencies between them have been captured in the sentence embeddings.

\section{Surface tasks: word content and order}

Though our ultimate interest is in abstract meaning information, part of the goal of these experiments is to get a clear picture of the information currently captured by existing systems. For this reason, we include the content and order experiments as performed by~\newcite{adi2016fine}, to see how encoding of these surface variables compares to encoding of meaning information---and to compare with the results of~\newcite{adi2016fine} after the more rigorous controls used in our datasets.  

We structure these tasks to be maximally parallel with our meaning tasks. To this end, we have two content tasks: one-probe (``\textbf{Content1Probe}'') and two-probe (``\textbf{Content2Probe}''), with the one-probe task using verb probes as in the negation task, and two-probe using noun-verb probe pairs, as in the semantic role task. Similarly, for the order task (``\textbf{Order}'') we use only noun-verb pairs. The order task is thus formulated as ``Given representation $\mathbf{n}$ of probe noun $n$, representation $\mathbf{v}$ of probe verb $v$, and embedding $\mathbf{s}$ of sentence $s$ (with $s$ containing both $n$ and $v$), does $n$ occur before $v$ in $s$?''. The two-word content task is formulated as ``Given representation $\mathbf{n}$ of probe noun $n$, representation $\mathbf{v}$ of probe verb $v$, and embedding $\mathbf{s}$ of sentence $s$, do both $n$ and $v$ occur in $s$?'', and the one-word content task is formulated as ``Given representation $\mathbf{v}$ of probe verb $v$, and embedding $\mathbf{s}$ of sentence $s$, does $v$ occur in $s$?'' 


\section{Classification experiments}

To demonstrate the utility of our analysis, we use it to test several existing sentence composition models. Following~\newcite{adi2016fine}, for our classifier we use a multi-layer perceptron with ReLU activations and a single hidden layer matching the input size. For each of the above tasks we construct train/test sets consisting of 4000 training items and 1000 test items.\footnote{See footnote 1 for link to all classification datasets used in these experiments.} No tuning is necessary, as the hyperparameters of hidden layer number and size are fixed in accordance with the architecture used by~\newcite{adi2016fine}.

It is important to note that the training of the classifier, which uses the 4000 items mentioned above, is to be distinguished from the training of the sentence embedding methods. The sentence embedding models are pre-trained on separate corpora, as described below, such that they map sentence inputs to embeddings.  Once these models are trained, they are used to produce the 4000 sentence embeddings that will serve as training input to the classifier (and the 1000 sentence embeddings used for testing).

Our use of a relatively simple classifier with a single hidden layer builds on the precedent not only of~\newcite{adi2016fine}, but also of related methods in neuroscience, which in fact typically use linear classifiers (an option that we could not employ due to our use of the variable probes). An important reason for use of simpler classifiers is to test for \emph{straightforward} extractability of information from embeddings---if a complex classifier is necessary in order to extract the information of interest, then this calls into question the extent to which we might consider this information to be ``captured'' in the embeddings, as opposed to the information being somehow reconstructable from the embeddings' encoding of other information. That said, the question of how the complexity of the classifier relates to the encoding of the target information in these sentence embeddings is an interesting issue for future work.

For each experiment, we also run two corresponding experiments, in which random vectors are used in place of the sentence vectors and the probes, respectively. This serves as an additional check for biases in the datasets, to ensure that neither the sentence vectors nor the probe vectors alone are sufficient to perform above chance on the tasks. For all tasks, these random vectors produce chance performance. 

\subsection{Sentence encoding models}

We test a number of composition models on these classification tasks. These models represent a range of influential current models designed to produce task-general sentence embeddings. They employ a number of different architectures and objectives, and have shown reasonable success on existing metrics \cite{hill2016learning,conneau2017supervised}.  

All sentence embeddings used are of 2400 dimensions. Because our pre-trained models (SDAE, Skip-Thought) are trained on the Toronto Books Corpus~\cite{zhu2015aligning}, we use this as our default training corpus, except when other supervised training data is required (as in the case of InferSent). Before sentence generation, the chosen vocabulary was checked against the training corpora to ensure that no words were out-of-vocabulary (or below a count of 50).

\paragraph{BOW averaging} 
Our first sentence embedding model (\textbf{``BOW''}) is a BOW averaging model, for which we use the skip-gram architecture of the word2vec model \cite{mikolov2013efficient} to learn word embeddings. As discussed above, the BOW model serves primarily as a sanity check for our purposes, but it is important to note that this model has had competitive results on various tasks, and is taken seriously as a sentence representation method for many purposes \cite{wieting2015towards,arora2016simple}. 

\paragraph{Sequential Denoising Autoencoder}
Our second model (\textbf{``SDAE''}) is an autoencoder variant from \newcite{hill2016learning} for unsupervised learning of sentence embeddings.  The model uses an LSTM-based encoder-decoder framework, and is trained to reconstruct input sentences from their vector representations (last hidden state of encoding LSTM) despite noise applied to the input sentence. We use a pre-trained model provided by the authors. This model has the advantage of an unsupervised objective and no need for sequential sentence data, and it shows competitive performance on a number of evaluations. 

\paragraph{Skip-Thought Embeddings}
Our next two models are variants of the Skip-Thought model \cite{kiros2015skip}, in which sentences are encoded with gated recurrent units (GRUs), with an objective of using the current sentence representation to predict the immediately preceding and following sentences. Following the model's authors, we use both the uni-skip (\textbf{``ST-UNI''}) and bi-skip (\textbf{``ST-BI''}) variants: uni-skip consists of an encoding based on a forward pass of the sentence, while bi-skip consists of a concatenation of encodings of the forward and backward passes of the sentence (each of 1200 dimensions, for 2400 total). We use the publicly available pre-trained Skip-Thought model for both of these variants.\footnote{\url{https://github.com/ryankiros/skip-thoughts}}

Skip-Thought sentence embeddings have been used as pre-trained embeddings for a variety of tasks. They have proven to be generally effective for supervised tasks and passable for unsupervised tasks \cite{hill2016learning,triantafillou2016towards,wieting2015towards}. Like the SDAE model, the Skip-Thought model is able to use unsupervised learning, though it requires sequential sentence data. However, more than the SDAE model, the Skip-Thought model uses an objective intended to capture semantic and syntactic properties, under the authors' assumption that prediction of adjacent sentences will encourage more syntactically and semantically similar sentences to map to similar embeddings.

\paragraph{InferSent}
Our final model is the InferSent model~\cite{conneau2017supervised}, which uses multi-layer BiLSTM encoders with max pooling on the hidden states of the last layer to produce vector representations of the sentences. This model is trained with a natural language inference (NLI) objective, and for this reason we train it on the SNLI dataset~\cite{bowman2015large}.

The InferSent model is intended to produce ``universal'' sentence representations, and has been shown to outperform unsupervised methods like Skip-Thought on a number of tasks~\cite{conneau2017supervised}. More generally, the NLI objective is believed to encourage learning of compositional meaning information, given that inference of entailment relations should require access to meaning information.


\subsection{Results and Discussion}

\begin{table*}
\centering
\begin{tabular}{cccccc|}
\cline{2-6}
& \multicolumn{5}{| c |}{Accuracy}  \\ 
&  \multicolumn{1}{| c}{Content1Probe}
&  \multicolumn{1}{c}{Content2Probe}
&  \multicolumn{1}{c}{Order}
&  \multicolumn{1}{c}{SemRole}
&  \multicolumn{1}{c |}{Negation}
\\ \hline
\multicolumn{1}{| c |}{BOW}& $100.0$& $97.1$& $55.0$& $51.3$& $50.9$\\
\multicolumn{1}{| c |}{SDAE}& $100.0$& $79.8$& $92.9$& $63.7$& $99.0$\\
\multicolumn{1}{| c |}{ST-UNI}& $100.0$& $88.1$& $93.2$& $62.3$& $96.6$\\
\multicolumn{1}{| c |}{ST-BI}& $96.6$& $79.4$& $88.7$& $63.2$& $74.7$\\
\multicolumn{1}{| c |}{InferSent}& $100.0$& $70.1$& $86.4$& $50.1$& $97.2$\\
\hline
\end{tabular}\caption{Classification results} \label{tab:results}
\end{table*}

Table~\ref{tab:results} shows the accuracy of the different models' sentence embeddings on our classification tasks. 

The first thing to note is that our BOW control allows us to confirm nearly complete lexical balance in the sentence sets: the averaged word embeddings perform roughly at chance on all but the content tasks.\footnote{The slightly higher accuracy on the order task is most likely the result of a very slight bias due to our use of only noun-verb order probe pairs for the sake of matching the SemRole task.}  
By contrast, BOW performs with near-perfect accuracy on the content tasks, lending support to the intuitive conclusion: the one thing that BOW \emph{does} encode is word content. The quality of performance of the BOW model on this task exceeds that reported by \newcite{adi2016fine}---we speculate that this may be due to our use of a smaller vocabulary to facilitate the learning of the mapping from one-hot probes.

While BOW has very high performance on two-probe word content, SDAE, ST-UNI, ST-BI and InferSent have much lower accuracy (albeit still far above chance), suggesting that some detail with respect to word content is sacrificed from these representations in favor of other information types. This is exemplified by the order task, on which all non-BOW models show significantly higher accuracy than on the word content tasks,  supporting the intuitive conclusion that such sequence-based models retain information about relative word position. This result is generally consistent with the \newcite{adi2016fine} result, but due to the additional control that brings BOW roughly to chance, we can conclude with greater confidence that the performance on this task pertains to order information in the source sentence itself. 

Turning to our meaning information tasks, we see that with the exception of ST-BI, the sequence models perform surprisingly well on the negation task, despite the fact that this task cannot be solved simply by detecting adjacency between negation and the verb (due to our insertion of adverbs). Instead, we speculate that these sequence models may be picking up on the utility of establishing a dependency between negation and the \emph{next} verb, even in the face of intervening words. This is not a complete solution to the problem of representing the meaning and dependencies of negation, but it is a useful step in that direction, and suggests that models may be sensitive to some of the behaviors of negation. 

Interestingly, ST-BI shows markedly weaker performance on the negation task. We see two potential reasons for this. First, it may be due to the reduced dimensionality of each of the two concatenated encodings (recall that ST-BI involves concatenating 1200-dimensional encodings of the forward and backward passes). Second, the reduced performance could be influenced by the inclusion of the backward pass: while the forward pass can leverage the strategy of linking negation to the next verb, the backward pass cannot use this strategy because it will encounter the relevant verb before encountering the negation. 

Turning to the semantic role task, we see a stark contrast with the high performance for the negation task. InferSent performs squarely at chance, suggesting that it retains as little compositional semantic role information as does BOW. SDAE, ST-UNI and ST-BI perform modestly above chance on the semantic role task at 62-63\% accuracy, suggesting that they may provide some amount of abstract role information---but no model shows any substantial ability to capture semantic role systematically.

These results accomplish two things. First, they lend credence to this method as a means of gaining insight into the information captured by current models. Second, they give us a sense of the current capacity of sequence-based models to capture compositional meaning information. The picture that emerges is that sequence models are able to make non-trivial headway in handling negation, presumably based on a sequential strategy of linking negation to the next verb---but that these sequence models fall significantly short when it comes to capturing semantic role compositionally. Another point that emerges from these results is that despite the fairly substantial differences in architecture, objective, and training of these models, capacity to capture the compositional information is fairly similar across models, suggesting that these distinct design decisions are not having a very significant impact on compositional meaning extraction. We plan to test more substantially distinct models, like those with explicit incorporation of syntactic structure~\cite{bowman2016fast,dyer2016recurrent,socher2013recursive} in future work. 

\section{Related work}
This work relates closely to a growing effort to increase interpretability of neural network models in NLP---including use of visualization to analyze what neural networks learn~\cite{li2015visualizing,kadar2016representation}, efforts to increase interpretability by generating explanations of model predictions~\cite{ribeiro2016should,lei2016rationalizing,li2016understanding}, and work submitting adversarial examples to systems in order to identify weaknesses \cite{zhao2017generating,jia2017adversarial,ettinger2017towards}. 

Methodologically the most closely related work is that of~\newcite{adi2016fine}, which uses classification tasks to probe for information in sentence embeddings. As discussed above, we depart from that work in targeting deeper and more linguistically-motivated aspects of sentence meaning, and we incorporate careful controls of our datasets to ensure elimination of bias in the results. 

Our focus on assessing linguistically-motivated information relates to work on evaluations that aim for fine-grained analysis of systems' linguistic capacities~\cite{Rim:Cla:Ste:09,Ben:Fli:Oep:11,marelli2014sick}. The present work contributes to this effort with new tasks that assess composition \emph{per se}, and that do so in a highly targeted manner via careful controls. Our use of synthetically generated data to achieve this level of control relates to work like that of \newcite{weston2015towards}, which introduces synthetic question-answering tasks for evaluating the capacity of systems to reason with natural language input.

Our examination of the capacity of neural sequence models to identify abstract relations in sentence representations also relates to work by \newcite{linzen2016assessing}, who explore whether LSTMs can learn syntactic dependencies, as well as \newcite{williams2017learning}, who investigate the extent to which parsers that are learned based on a semantic objective produce conventional syntax. 


Finally, importantly related work is that concerned specifically with testing systematic composition. \newcite{lake2017still} investigate the capacity of RNNs to perform zero-shot generalization using composition, and \newcite{dasgupta2018evaluating} construct an entailment dataset with balanced lexical content in order to target composition more effectively. We contribute to this line of inquiry by establishing an analysis method that can take output embeddings from sentence composition models and query them directly for specific types of information to be expected in properly compositional sentence representations.

\section{Conclusions and future directions}
We have presented an analysis method and accompanying generation system designed to address the problem of assessing compositional meaning content in sentence vector representations. We make the datasets for these tasks, as well as the generation system used to create them, available for public use to facilitate broader testing of composition models. We have also presented the results of applying this method for analysis of a number of current sentence composition models, demonstrating the capacity of the method to derive meaningful information about what is captured in these models' outputs. 

Having established a means of analyzing compositional meaning information in sentence embeddings, in future work we plan to apply this system to identify more precisely which design decisions lead to effective capturing of meaning information, in order to guide system improvement. As part of this effort, we will expand to more comprehensive testing of a diverse range of sentence embedding systems~\cite{bowman2016fast,subramanian2018learning}. We also plan to investigate the potential of our generation system to create not just evaluation data, but training data---given that it allows us to produce large, meaning-annotated corpora. Finally, we plan to expand beyond semantic role and negation in the set of information types targeted by our method, in order to establish more comprehensive coverage of meaning information that can be assessed by this analysis system. 

\section*{Acknowledgements}

Devin Ganey contributed code interfacing the generation system and the XTAG database. The work described in this paper benefited from discussions with Alexander Williams, Marine Carpuat, and Hal Daum\'e III, and from helpful comments by Ido Dagan, Jason Eisner, Chris Dyer, audience members at RepEval 2016, members of the UMD CLIP and CNL labs, as well as anonymous reviewers. This work was supported in part by an NSF Graduate Research Fellowship to Allyson Ettinger under Grant No.~DGE 1322106, and by NSF NRT Grant DGE-1449815. Any opinions, findings, and conclusions or recommendations expressed are those of the authors and do not necessarily reflect the views of the NSF.

\bibliographystyle{acl}
\bibliography{compevalnaacl}

%
%
%
%
%
%

\end{document}